%% file: neurips_2026.tex
\documentclass{article}

\usepackage[preprint]{neurips_2026}

\usepackage[utf8]{inputenc} 
\usepackage[T1]{fontenc}    
\usepackage{hyperref}       
\usepackage{url}            
\usepackage{booktabs}       
\usepackage{amsfonts}       
\usepackage{nicefrac}       
\usepackage{microtype}      
\usepackage{xcolor}         
\usepackage{enumitem}
\usepackage{graphicx}
\usepackage{tabularx}
\usepackage{amsthm}
\usepackage{tikz}
\usetikzlibrary{arrows.meta, positioning, calc, fit, backgrounds}
\newtheorem{proposition}{Proposition}
\newtheorem{theorem}{Theorem}
\newtheorem{lemma}{Lemma}
\title{Q-Interference:  Memory-Efficient Phase-Aware Quantum-Inspired Attention}

\author{
Emama Nahid \\
Kennesaw State University \\
Marietta, GA, USA \\
\texttt{enahid@students.kennesaw.edu} \\
\And
Tahmid Imtiaz Imu \\
Kennesaw State University \\
Marietta, GA, USA \\
\texttt{timu1@students.kennesaw.edu} \\
\And
Huayue Gu \\
Kennesaw State University \\
Marietta, GA, USA \\
\texttt{hgu2@kennesaw.edu} \\
\AND
Liran Ma \\
Miami University \\
Oxford, OH, USA \\
\texttt{mal18@MiamiOH.edu} \\
\And
Zhipeng Cai \\
Georgia State University \\
Atlanta, GA, USA \\
\texttt{zcai@gsu.edu} \\
\And
Honghui Xu \\
Kennesaw State University \\
Marietta, GA, USA \\
\texttt{hxu10@kennesaw.edu} \\
}

\begin{document}

\maketitle

\begin{abstract}
GPT attention measures token compatibility through dot-product similarity. This mechanism is simple, effective, and memory-efficient. But it does not explicitly model whether strong token features should reinforce or suppress one another. We introduce Q-Interference, a fully classical quantum-inspired attention mechanism for autoregressive language modeling that augments each query and key feature with an amplitude and a learned phase. The resulting attention score is phase-aware which aligned phases contribute constructively while conflicting phases contribute destructively. Although Q-Interference yields a richer interaction rule than similarity alone, a naive implementation of Q-Interference requires a large token-pair-feature interaction tensor, making it memory-intensive and often impractical. To address this limitation, we propose an exact trigonometric factorization that computes the same score using two standard matrix multiplications avoiding materialization of the large intermediate tensor. Q-Interference fits directly into a Transformer block in GPT and leaves the remainder of the model architecture and next-token prediction objective unchanged. Experiments on public benchmark datasets and baseline models show that the proposed reformulation trains stably in a controlled GPT-style setting and provides a consistent memory advantage over naive phase-aware interference attention. These results support the specific contribution of this work: an exact memory-efficient reformulation that makes phase-aware interference attention practical within a standard GPT pipeline. Our code is available at \url{https://anonymous.4open.science/r/Q-Interference-Memory-Efficient-Quantum-Inspired-Attention-BDF9}.
\end{abstract}

\section{Introduction}

Transformer based GPT models have become a foundation of modern language processing because they can represent context across token sequences \cite{attentionIsAllYouNeed,bert2019}. A token is a basic text unit. Self attention is the mechanism that lets each token compare itself with other tokens in the same sequence. GPT style autoregressive models use causal self attention where each position attends only to previous positions to predict the next token \cite{fewShotLearner2020}. However, standard attention usually measures token compatibility through a dot product which mainly reflects feature similarity. This view may miss cases where strong features should support or suppress each other depending on context. 

\noindent This challenge becomes more important in long context language modeling \cite{longcontext2019}. Practical systems such as document question answering, scientific text modeling and retrieval augmented generation often depend on information spread across many tokens \cite{longformer2020,RAG}. These settings require a model to connect nearby words with distant evidence. Standard dense attention lets each token compare with many earlier tokens. So its memory cost grows with sequence length \cite{flashAttention,longformer2020}. During autoregressive inference, the key value cache stores past keys and values and also grows as more tokens are generated \cite{qubitCache2026,memoryManagement2023}. Consequently, memory efficiency is not only a systems concern but also a condition for making richer attention rules practical at scale.

\noindent Prior research has improved attention and language model efficiency through better memory use, cheaper sequence computation and richer interaction structure \cite{surveyTransformer2022}. 
On the efficiency side, standard attention has been accelerated or approximated through memory aware exact computation, low rank structure, random feature estimation, structural routing and sparse access patterns \cite{flashAttention,linFormer2020,performer2020,reformer2024,bigbird2020}. 
These advances make long sequence modeling more practical, but they largely keep token compatibility tied to the standard query key similarity view. 
On the modeling side, quantum and quantum inspired attention has introduced richer ways to represent token relationships through state based interaction, circuit motivated computation and quantum style similarity \cite{quantumSelfAttentionText2024,quantumMixedSelfAttention2025,hybridQuantumTransformer2025,quantumInspiredSelfAttention2026}. 
Quantum inspired efficiency has also been used to reduce fine tuning cost and key value cache storage, but these contributions operate outside the internal score computation that creates the phase aware interaction tensor \cite{quanTa2024,qubitCache2026}.
Together, these studies suggest that attention can be made both more efficient and more expressive. 
However, they do not address the specific memory overhead created when phase aware pairwise feature interactions are implemented directly inside GPT style autoregressive language modeling. 
This leaves a gap for an exact tensor memory-efficient reformulation that preserves constructive and destructive phase interactions while avoiding the large intermediate interaction tensor.

\noindent We introduce Q-Interference to address this gap. It is a fully classical quantum inspired attention mechanism for autoregressive language modeling. The method augments each query and key feature with an amplitude and a learned phase. The amplitude controls the strength of a feature, while the phase controls how that feature interacts with another token. Aligned phases create constructive interaction, whereas conflicting phases create destructive interaction. A direct implementation of this idea requires a large token pair feature tensor. So we derive an exact factorization that computes the same score using two standard matrix multiplications. This makes the phase-aware computation memory efficient with respect to its additional interaction cost, while preserving the GPT backbone and the next token prediction objective. Our contributions are as follows.
\begin{itemize}[leftmargin=*, itemsep=1pt, topsep=2pt]
    \item We introduce Q-Interference, a phase aware quantum inspired attention mechanism for GPT style language modeling. It uses amplitude and learned phase to model constructive and destructive token interactions.
    \item An exact memory efficient reformulation is derived for phase aware interference attention. The reformulation computes the same attention score with two standard matrix multiplications. 
    \item Q-Interference is evaluated in a controlled GPT style setup. The backbone and next token objective remain unchanged so that the effect of the attention rule can be isolated.
\end{itemize}

\begin{figure*}[t]
    \centering
    \includegraphics[width=\textwidth]{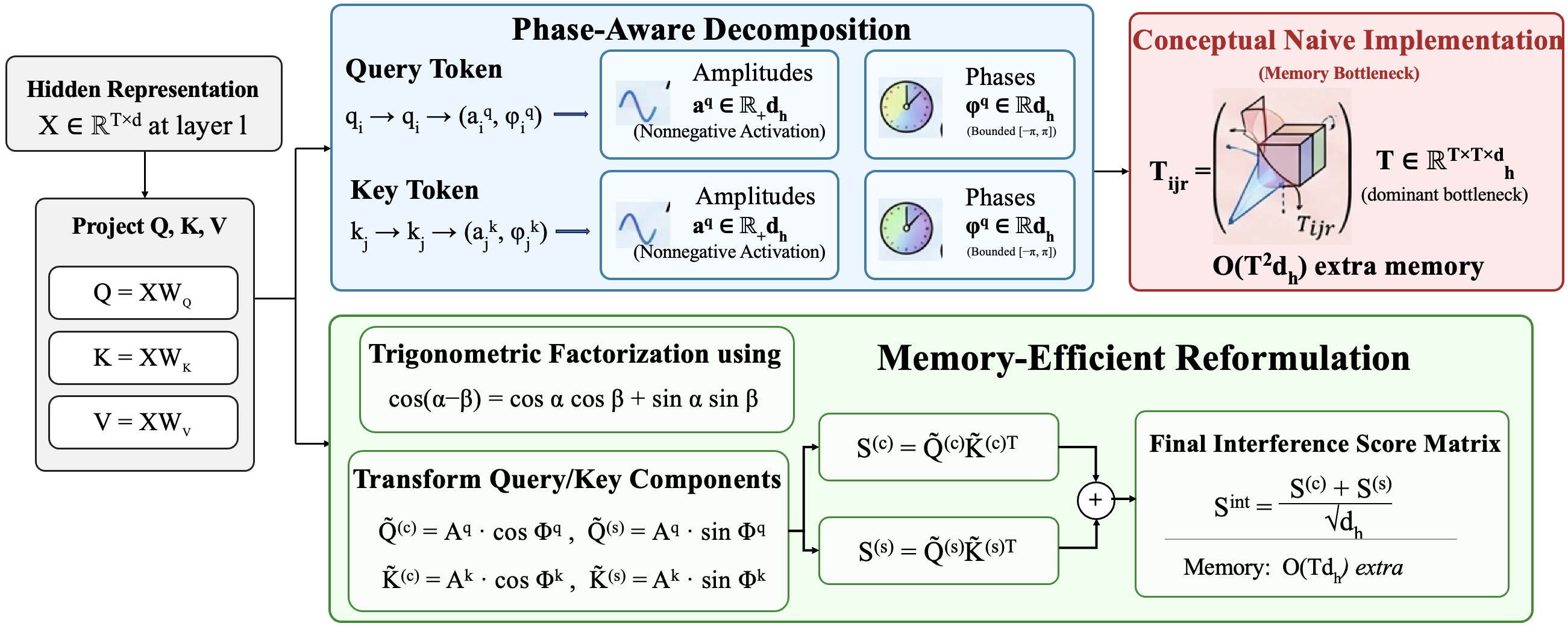}
    \caption{Q-Interference methodology flowchart.}
    \label{fig:methodology}
\end{figure*}

\section{Q-Interference}
We introduce Q-Interference, a quantum-inspired variant of GPT in which the standard dot-product attention score is replaced by a phase-aware interference score. The model remains entirely classical and is trained on standard GPU hardware. The aim is not to claim quantum advantage, but to incorporate a useful structural idea from wave-like interactions into autoregressive language modeling. Notation and proof assumptions are collected in Appendix~\ref{app:notation}.
 The design is intentionally minimal. We preserve the standard GPT backbone, including token and positional embeddings, residual connections, layer normalization, feed-forward blocks, and the next-token prediction objective. As shown in Fig.~\ref{fig:methodology}, the only modification lies in the attention scoring rule. This keeps the comparison against a standard GPT baseline controlled and makes the source of improvement easier to interpret .
Let $X \in \mathbb{R}^{T \times d}$ denote the hidden sequence representation at a given layer, where $T$ is the sequence length and $d$ is the model dimension. Standard self-attention computes the query, key, and value projections as $Q = XW_Q$, $K = XW_K$, and $V = XW_V$.
where $W_Q, W_K, W_V \in \mathbb{R}^{d \times d_h}$ are learned projections for a head of dimension $d_h$. The conventional attention score between tokens $i$ and $j$ is $s_{ij}^{\mathrm{std}} = \frac{q_i^\top k_j}{\sqrt{d_h}}$.
After applying the causal mask $M$, the normalized attention weights are
\begin{equation}
\alpha_{ij} = \frac{\exp\!\left(s_{ij}^{\mathrm{std}} + M_{ij}\right)}
{\sum_{m \le i} \exp\!\left(s_{im}^{\mathrm{std}} + M_{im}\right)}
\end{equation} and the output at position $i$ is $o_i = \sum_{j \le i} \alpha_{ij} v_j$.

This reformulation is effective, but it assumes that token compatibility is sufficiently captured by magnitude-based similarity. Our method relaxes that assumption by introducing an explicit phase component into token interaction.

\noindent
\textbf{Phase-Aware Interference Attention:}
Standard self-attention measures token compatibility through the query-key dot product. Although effective, this score is mainly magnitude-driven and treats strong feature alignment as supportive interaction. In language, however, token relationships can be either reinforcing or suppressive depending on context. To model this more explicitly, we introduce a quantum-inspired attention mechanism that combines feature strength with relative phase alignment.

Our goal is not to claim quantum advantage, but to use a simple wave-inspired principle: aligned signals interact constructively, while misaligned signals interact destructively. This gives the attention score a more expressive way to capture both supportive and conflicting token relationships.

Instead of using real-valued query and key vectors alone, we represent each projected feature using an amplitude-phase decomposition: 
\begin{equation}
    q_i \rightarrow \left(a_i^q, \phi_i^q\right), k_j \rightarrow \left(a_j^k, \phi_j^k\right),
\end{equation} where $a_i^q, a_j^k \in \mathbb{R}_+^{d_h}$ are nonnegative amplitudes and $\phi_i^q, \phi_j^k \in \mathbb{R}^{d_h}$ are learned phases. In practice, amplitudes are produced through a nonnegative activation, while phases are constrained to a bounded interval such as $[-\pi, \pi]$ for numerical stability.
This decomposition gives a simple interpretation. The amplitude controls how strongly a feature participates in attention, while the phase controls how that feature interacts with another one. Two features with aligned phase reinforce each other, whereas two features with conflicting phase suppress one another. In this way, token interaction is no longer determined only by feature strength, but also by relative phase alignment.
Given a query token $i$ and a key token $j$, we define the interference-based attention score as
\begin{equation}
s_{ij}^{\mathrm{int}} = \frac{1}{\sqrt{d_h}} \sum_{r=1}^{d_h}
a_{i,r}^q\, a_{j,r}^k \cos\!\left(\phi_{i,r}^q - \phi_{j,r}^k\right)
\end{equation}

This is the main departure from standard attention. Each feature contributes not only according to its amplitude, but also according to the cosine of its phase difference. When the phases are close, the cosine term is positive and the interaction is strengthened. When the phases disagree, the cosine term decreases or becomes negative, suppressing the interaction. As a result, the model can distinguish constructive and destructive relationships between tokens rather than treating all large-magnitude alignments as equally supportive. An algebraic derivation of the phase-aware interference score is provided in Appendix~\ref{app:dev_score}.

The trade-off is that this richer interaction rule is more expensive than classical dot-product attention when implemented naively. In particular, directly computing phase-aware pairwise interactions introduces an additional intermediate structure over token pairs and feature dimensions, which significantly increases memory usage. This makes the method more expressive, but also potentially less practical at scale. To address this issue, we later introduce a memory-efficient factorization that preserves the same interference score while avoiding the explicit construction of the large intermediate tensor.
Once the score is computed, the rest of the attention pipeline remains unchanged: $\alpha_{ij}
=
\frac{\exp\!\left(s_{ij}^{\mathrm{int}} + M_{ij}\right)}
{\sum_{m \le i} \exp\!\left(s_{im}^{\mathrm{int}} + M_{im}\right)}$, 
$o_i = \sum_{j \le i} \alpha_{ij} v_j$.
Thus, the proposal changes only the compatibility function inside self-attention, while preserving the standard GPT transformer  workflow.

\noindent
\textbf{Memory-Efficient Reformulation:}
The phase-aware attention defined above gives a richer compatibility score than standard dot-product attention by allowing token interactions to be reinforced or weakened through relative phase alignment. The drawback is that a direct implementation requires pairwise phase interactions across all token pairs and feature dimensions, which introduces a large intermediate tensor and high memory cost. Our method addresses this issue by retaining the phase-aware score while avoiding the main memory bottleneck of the naive computation.
A direct implementation of the interference score is computationally impractical. The naive formulation constructs an interaction tensor over token pairs and feature dimensions: $\mathcal{T}_{ijr} = a_{i,r}^q\, a_{j,r}^k \cos\!\left(\phi_{i,r}^q - \phi_{j,r}^k\right)$.
The final score is then obtained by summing over the feature index $r$: $s_{ij}^{\mathrm{int}} = \frac{1}{\sqrt{d_h}} \sum_{r=1}^{d_h} \mathcal{T}_{ijr}$.

However, $\mathcal{T} \in \mathbb{R}^{T \times T \times d_h}$ introduces a large intermediate tensor that quickly becomes the dominant memory bottleneck. This is exactly the inefficiency we aim to remove.
Our key technical contribution is an exact trigonometric factorization of the interference score. Using the identity: $\cos(\alpha - \beta) = \cos \alpha \cos \beta + \sin \alpha \sin \beta$, we rewrite the score as
\begin{equation}
    s_{ij}^{\mathrm{int}}
=
\frac{1}{\sqrt{d_h}}
\sum_{r=1}^{d_h}
\left(a_{i,r}^q \cos \phi_{i,r}^q\right)
\left(a_{j,r}^k \cos \phi_{j,r}^k\right)
+
\frac{1}{\sqrt{d_h}}
\sum_{r=1}^{d_h}
\left(a_{i,r}^q \sin \phi_{i,r}^q\right)
\left(a_{j,r}^k \sin \phi_{j,r}^k\right).
\end{equation}

Then, we can define transformed query and key components:
\[
\tilde{q}_{i,r}^{(c)} = a_{i,r}^q \cos \phi_{i,r}^q,
\qquad
\tilde{q}_{i,r}^{(s)} = a_{i,r}^q \sin \phi_{i,r}^q,
\]
\[
\tilde{k}_{j,r}^{(c)} = a_{j,r}^k \cos \phi_{j,r}^k,
\qquad
\tilde{k}_{j,r}^{(s)} = a_{j,r}^k \sin \phi_{j,r}^k.
\]

Substituting these definitions yields
$s_{ij}^{\mathrm{int}}
=
\frac{1}{\sqrt{d_h}}
\left(
\tilde{q}_i^{(c)\top}\tilde{k}_j^{(c)}
+
\tilde{q}_i^{(s)\top}\tilde{k}_j^{(s)}
\right).
$ Finally, the full score matrix becomes
\begin{equation}
    S^{\mathrm{int}}
=
\frac{
\tilde{Q}^{(c)} \tilde{K}^{(c)\top}
+
\tilde{Q}^{(s)} \tilde{K}^{(s)\top}
}{\sqrt{d_h}}.
\end{equation}
This reformulation is exact and introduces no approximation. It avoids explicitly forming the  $T \times T \times d_h$ interaction tensor by rewriting the phase-aware score as two standard matrix multiplications and an addition. Consequently, the extra memory cost drops from $\mathcal{O}(T^2 d_h)$ in the naive form to $\mathcal{O}(T d_h)$ in the factorized form. The final score matrix remains standard attention-sized, so our claim is not to eliminate all quadratic costs of dense attention, but to remove the additional memory overhead introduced by naive phase-aware interaction. Formal proofs of exact score equivalence and phase-specific memory accounting are provided in Appendices~\ref{app:ex_factor} and~\ref{app:mem_cost_comp}.

\noindent
\textbf{Integration and Training:}
The proposed score is inserted into a standard GPT block without changing the rest of the model architecture. Let $H^{(\ell)}$ denote the hidden representation at layer $\ell$. The layer update follows the usual residual form:
\begin{equation}
    \hat{H}^{(\ell)} = H^{(\ell)} + \mathrm{MHA}_{\mathrm{int}}\!\left(\mathrm{LN}\!\left(H^{(\ell)}\right)\right),
\end{equation}
\begin{equation}
    H^{(\ell+1)} = \hat{H}^{(\ell)} + \mathrm{MLP}\!\left(\mathrm{LN}\!\left(\hat{H}^{(\ell)}\right)\right).
\end{equation}

Here, $\mathrm{MHA}_{\mathrm{int}}$ denotes multi-head attention using the proposed interference score. All remaining components are inherited directly from the baseline GPT architecture. This keeps the architectural intervention narrow and makes the comparison with the original model more reliable. A proof of compatibility with the standard causal GPT attention interface is provided in Appendix~\ref{app:comp_cgpt}.

Training is performed with the standard autoregressive language modeling objective. Given a token sequence $(x_1, x_2, \ldots, x_T)$, we minimize
\begin{equation}
    \mathcal{L}_{\mathrm{LM}} = - \sum_{t=1}^{T-1} \log p_{\theta}(x_{t+1} \mid x_{\leq t}).
\end{equation}
No auxiliary objective is introduced. This keeps the setup identical to standard autoregressive GPT training and isolates the effect of the proposed attention mechanism.

\section{Experiment and Evaluation}

We evaluate Q-Interference in a controlled GPT-style autoregressive language modeling setting, where the standard scaled dot-product attention score is replaced by the proposed phase-aware interference score while the rest of the GPT transformer backbone is kept unchanged. This design keeps token embeddings, positional embeddings, residual connections, layer normalization, feed-forward blocks, and the next-token prediction objective fixed, so that observed differences can be attributed primarily to the attention mechanism itself. Our experiments are designed to answer three questions. \textbf{RQ1:} whether Q-Interference can be trained successfully inside a standard GPT pipeline? \textbf{RQ2:} whether the exact factorization removes the extra memory overhead introduced by naive phase-aware interaction? \textbf{RQ3:} how the resulting model compares with standard and pre-trained GPT transformer baselines in terms of language-modeling quality and efficiency?

\subsection{Experiment Setup}

\noindent
\textbf{Dataset and Preprocessing:}
We use four language modeling datasets: WikiText-103\cite{wikitext}, TinyStories \cite{tinystories2024}, pile-10k \cite{pile} , and small-C4 \cite{smallc4}. WikiText-103 serves as the main controlled benchmark, while the other three datasets are used to evaluate cross-dataset behavior under different data regimes. We use the GPT-2 tokenizer and segment each corpus into fixed-length sequences with context length 512.

\noindent
\textbf{Baselines:} 
We compare six models. We first consider two baselines: a standard GPT baseline with conventional causal self-attention and a Q-GPT: quantum inspired baseline adopted from a previously published quantum-inspired GPT study \cite{gptOnQuantumComputer2024}, which serves as a reference model architectural comparison and is not our contribution. We then include two widely used pre-trained decoder-only reference models, GPT-Neo-125M \cite{gptneo2021} and OPT-125M \cite{opt2022}. To clarify the evaluation design, Table~\ref{tab:baseline_rationale} summarizes the role of each baseline and why the selected set is sufficient for the scope of this study.

\begin{table*}[t]
\centering
\caption{Baseline selection rationale.}
\label{tab:baseline_rationale}
\vspace{1mm}
\small
\setlength{\tabcolsep}{4pt}
\renewcommand{\arraystretch}{1.10}
\begin{tabular}{p{2.7cm} p{3.2cm} p{7.5cm}}
\hline
\textbf{Model} & \textbf{Comparison role} & \textbf{Justification} \\
\hline
Baseline GPT 
& Standard internal control 
& Directly controlled reference for conventional causal dot-product attention under the same GPT-style training setup. \\
\hline
Q-GPT baseline 
& Quantum-inspired reference 
& Provides a published quantum-inspired GPT-style architectural comparison in this setting. \\
\hline
OPT-125M and GPT-Neo-125M  
& External pretrained reference 
& Gives context against an open autoregressive GPT-style model family. \\
\hline
\end{tabular}
\end{table*}

\noindent
\textbf{Training Configurations:} 
Finally, we evaluate two versions of our phase-aware model family: a naive interference model that directly computes the phase-aware interaction tensor, and the proposed Q-Interference model, which uses the exact memory-efficient reformulation. The naive model is used primarily for profiling and ablation because it is substantially more memory-intensive. For the main controlled comparison, the standard GPT baseline has approximately 124.0M parameters, while the matched Q-Interference model has approximately 123.7M parameters, using 12 layers, 12 heads, and model dimension 720. Unless otherwise stated, all models are trained with the standard autoregressive objective on NVIDIA Tesla V100-SXM2-32GB GPUs using mixed precision.

\noindent
\textbf{Evaluation Metrics:}
We report validation loss, test loss, and test perplexity as the primary language-modeling metrics. To evaluate practicality, we also report peak GPU memory usage, elapsed execution time, and sequence-length scaling behavior. Peak GPU memory was measured under fixed batch size, context length, numerical precision, and hardware; therefore, the reported peak memory values mainly reflect the computational behavior of the model architecture rather than the identity of the dataset itself. This is why peak memory can remain nearly constant across datasets for a given model. Our goal is not to eliminate the standard quadratic attention matrix, but to remove the additional memory cost introduced by naive phase-aware interaction while preserving the same interference-based scoring idea. Definitions of the reported metrics and model categories are summarized in Appendix~\ref{app:metrix_def}.

\subsection{Comparison Evaluation with Baselines}
In this section, Q-Interference is evaluated from two comparison perspectives. First, the proposed model is compared with the baselines across the evaluated datasets and it is compared against 125M pre-trained reference models.

\begin{figure*}[t]
    \centering
    \includegraphics[width=0.92\textwidth]{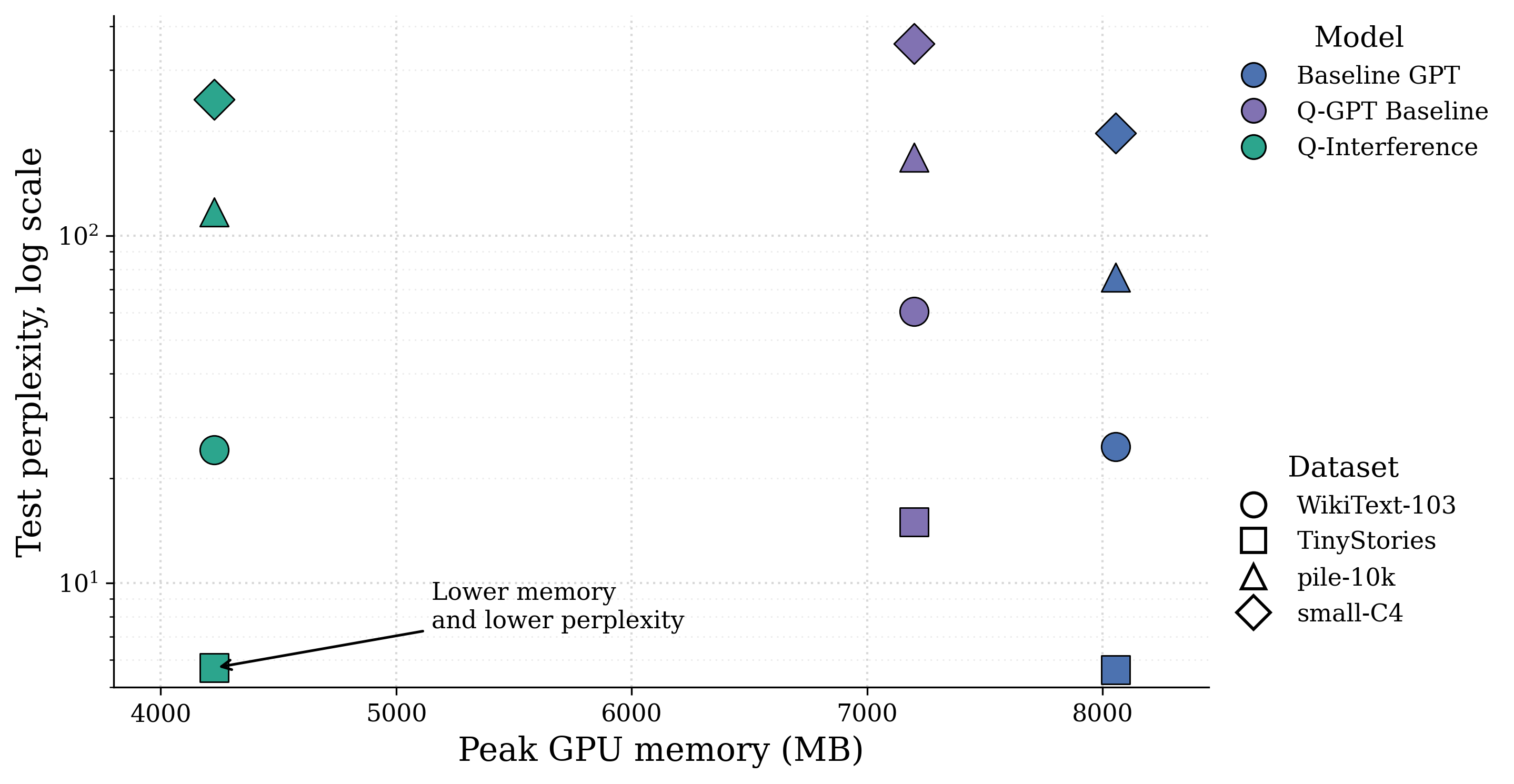}
    \caption{Quality-memory trade-off for the main model comparison.}
    \label{fig:quality_memory_tradeoff}
\end{figure*}

\noindent
\textbf{Comparison across datasets:}
We evaluate the standard GPT baseline, the proposed Q-Interference model, and the Q-GPT baseline across four datasets: WikiText-103, TinyStories, pile-10k, and small-C4. WikiText-103 serves as the main benchmark, while the remaining datasets are used to examine whether the observed behavior transfers across different language-modeling regimes. In all cases, the comparison is performed under the same overall GPT-style training pipeline so that the effect of the proposed phase-aware hybrid attention can be isolated while keeping the backbone architecture and next-token prediction objective unchanged. Table~\ref{tab:cross_dataset_results} shows a consistent practical advantage for Q-Interference together with mixed but competitive quality behavior. On the main benchmark, WikiText-103, Q-Interference is the strongest internal model, improving validation loss, test loss, and test perplexity over the standard GPT baseline while also reducing peak training GPU memory from 8055.76 MB to 4227.14 MB. A visual summary of the WikiText-103 comparison is provided in Appendix~\ref{app:wiki_result}. On TinyStories, it remains very close to the standard GPT baseline in test loss and perplexity while preserving the same memory advantage. On pile-10k and small-C4, the standard GPT baseline remains stronger in final test quality, but Q-Interference is still substantially more memory-efficient and remains clearly stronger than the Q-GPT baseline. Figure~\ref{fig:quality_memory_tradeoff} shows the relationship between test perplexity and peak GPU memory across datasets, highlighting that Q-Interference achieves the most favorable practical trade-off within the phase-aware model family. Overall, these results suggest that the proposed method is best understood not as a universal replacement for standard GPT, but as the strongest practical model within the phase-aware family, offering the most favorable quality-efficiency trade-off across the evaluated datasets. A supplementary visualization of memory behavior under the fixed profiling setup is provided in Appendix~\ref{app:data_memory}.

{\begin{table*}[t]
\centering
\small
\caption{Baselines vs Q-Interference comparison over benchmark dataset }
\label{tab:cross_dataset_results}
\resizebox{\textwidth}{!}{\begin{tabular}{llcccc}
\hline
Dataset & Model & Best Val Loss & Test Loss & Test PPL & Peak GPU Mem (MB) \\
\hline
WikiText-103 & Baseline GPT & 3.2036 & 3.2049 & 24.6534 & 8055.76 \\
WikiText-103 & Q-GPT baseline & 4.1142 & 4.1032 & 60.5367 & \ 7199.26 \\
WikiText-103 & Q-Interference (ours) & \textbf{3.1809} & \textbf{3.1852} & \textbf{24.1718} & \textbf{4227.14} \\
\hline
TinyStories & Baseline GPT & \textbf{1.7279} & \textbf{1.7263} & \textbf{5.6196} & 8055.76 \\
TinyStories & Q-GPT baseline & 2.7036 & 2.7082 & 15.0016 & 7199.26 \\
TinyStories & Q-Interference (ours) & 1.7344 & 1.7394 & 5.6941 & \textbf{4227.14} \\
\hline
pile-10k & Baseline GPT & 4.1993 & \textbf{4.3294} & \textbf{75.9008} & 8055.76 \\
pile-10k & Q-GPT baseline & 4.9723 & 5.1241 & 168.0210 & 7199.26 \\
pile-10k & Q-Interference (ours) & \textbf{4.0745} & 4.7616 & 116.9275 & \textbf{4227.14} \\
\hline
small-C4 & Baseline GPT & \textbf{5.3191} & \textbf{5.2833} & \textbf{197.0284} & 8055.76 \\
small-C4 & Q-GPT baseline & 5.9414 & 5.8769 & 356.7006 & 7199.26 \\
small-C4 & Q-Interference (ours) & 5.5224 & 5.5090 & 246.9012 & \textbf{4227.14} \\
\hline
\end{tabular} }
\end{table*}

\noindent
\textbf{Comparison with pre-trained reference models:}
To further contextualize our results, we compare Q-Interference with two widely used pretrained decoder-only reference models, GPT-Neo-125M and OPT-125M. Since these are not parameter-matched internal baselines, we treat them as pre-trained reference points rather than direct apples-to-apples competitors.
As shown in Table~\ref{tab:external_reference_results}, GPT-Neo-125M and OPT-125M achieve stronger final language-modeling quality than Q-Interference on all four datasets, with the gap being especially large on pile-10k and small-C4. At the same time, Q-Interference remains more memory-efficient on TinyStories, pile-10k, and small-C4, and is competitive with these models on WikiText-103 in terms of training memory. Thus, the results position Q-Interference not as a universal replacement for strong pretrained GPT transformers, but as the strongest practical phase-aware model in our experiments with a favorable efficiency profile. A supplementary visual comparison to the reference models is provided in Appendix~\ref{app:pretrain}.

\begin{table*}[t]
\centering
\small
\caption{Comparison to pre-trained reference models.}
\label{tab:external_reference_results}
\begin{tabular}{llccc}
\hline
Dataset & Model & Test Loss & Test PPL & Peak GPU Mem (MB) \\
\hline
WikiText-103 & GPT-Neo-125M & \textbf{2.9673} & \textbf{19.4386} & 4097.91 \\
WikiText-103 & OPT-125M & 3.0169 & 20.4278 & \textbf{4213.64} \\
WikiText-103 & Q-Interference (ours) & 3.1852 & 24.1718 & 4227.14 \\
\hline
TinyStories & GPT-Neo-125M & 1.5198 & 4.5713 & 9934.54 \\
TinyStories & OPT-125M & \textbf{1.4217} & \textbf{4.1443} & 6219.52 \\
TinyStories & Q-Interference (ours) & 1.7394 & 5.6941 & \textbf{4227.14} \\
\hline
pile-10k & GPT-Neo-125M & \textbf{2.3898} & \textbf{10.9116} & 9934.54 \\
pile-10k & OPT-125M & 2.7873 & 16.2376 & 6219.52 \\
pile-10k & Q-Interference (ours) & 4.7616 & 116.9275 & \textbf{4227.14} \\
\hline
small-C4 & GPT-Neo-125M & 3.5453 & 34.6512 & 9931.54 \\
small-C4 & OPT-125M & \textbf{3.4794} & \textbf{32.4405} & 6219.52 \\
small-C4 & Q-Interference (ours) & 5.5090 & 246.9012 & \textbf{4227.14} \\
\hline
\end{tabular}
\end{table*}

\noindent
Overall, the results show that Q-Interference is the strongest practical model within the proposed phase-aware family. It achieves the best internal matched result on WikiText-103 and offers the most favorable quality-memory trade-off across the custom models we study. While the gains are not universal across all datasets, the proposed design consistently remains more practical than the Q-GPT baseline.

\subsection{Ablation Study}

We design the ablation study to isolate the two main claims of this work. First, we test whether the memory-efficient reformulation is necessary for making phase-aware attention practical. Second, we test whether the phase component itself contributes to language-modeling quality inside the final hybrid design. Unlike the main results section, which emphasizes overall model comparison, the goal here is to separate the systems contribution from the modeling contribution.

\noindent
\textbf{Naive phase-aware attention versus memory-reduced Q-Interference:}
The first ablation compares the naive phase-aware implementation directly with the final Q-Interference model across the evaluated datasets. The naive model uses the same interference-based scoring idea, but computes it without the memory-efficient reformulation, and therefore serves as the most direct control for testing whether the proposed reformulation is necessary in practice.
As shown in Table~\ref{tab:ablation_naive_vs_ours}, the systems motivation of the paper is directly supported. On WikiText-103, the full matched naive run is impractical and reaches out-of-memory at context length 512, whereas Q-Interference remains fully trainable. On TinyStories, pile-10k, and small-C4, peak training memory is reduced from 12138.34 MB to 4227.14 MB, corresponding to a reduction of about 65\%. At the same time, stronger final test quality is achieved on the completed datasets. Although the naive model attains a slightly lower validation loss on pile-10k, it performs worse in final test loss and perplexity. Overall, these results indicate that the efficient reformulation is not a minor implementation detail but the key step that makes phase-aware attention practical.

\begin{table*}[t]
\centering
\caption{Naive phase-aware attention versus Q-Interference.}
\label{tab:ablation_naive_vs_ours}
\resizebox{\textwidth}{!}{\begin{tabular}{llcccc}
\hline
Dataset & Model & Best Val Loss & Test Loss & Test PPL & Peak GPU Mem (MB) \\
\hline
WikiText-103 & Naive interference & OOM & OOM & OOM & OOM \\
WikiText-103 & Q-Interference (ours) & \textbf{3.1809} & \textbf{3.1852} & \textbf{24.1718} & \textbf{4227.14} \\
\hline
TinyStories & Naive interference & 2.2972 & 2.3422 & 10.4037 & 12138.34 \\
TinyStories & Q-Interference (ours) & \textbf{1.7344} & \textbf{1.7394} & \textbf{5.6941} & \textbf{4227.14} \\
\hline
pile-10k & Naive interference & \textbf{4.0223} & 5.1086 & 165.4355 & 12138.34 \\
pile-10k & Q-Interference (ours) & 4.0745 & \textbf{4.7616} & \textbf{116.9275} & \textbf{4227.14} \\
\hline
small-C4 & Naive interference & 5.8665 & 5.8401 & 268.3050 & 12138.34 \\
small-C4 & Q-Interference (ours) & \textbf{5.5224} & \textbf{5.5090} & \textbf{246.9012} & \textbf{4227.14} \\
\hline
\end{tabular}}
\end{table*}

\noindent
\textbf{Effect of the phase component:}
The second ablation is designed to evaluate whether the phase term contributes meaningfully within the final hybrid design. For this purpose, the full Q-Interference model is compared with a phase-disabled control in which the same hybrid architecture and training setup are retained, but the learned phase contribution is removed. The corresponding results are reported in Table~\ref{tab:ablation_nophase}.
Overall, a small but consistent improvement in final test quality is observed when the phase term is retained. Across all four datasets, slightly better test loss and perplexity are achieved by the full model than by the phase-disabled variant, while only a modest difference in memory is observed. These results suggest that the main practical benefit of the method is provided by the memory-efficient reformulation, whereas an additional but smaller modeling gain is supplied by the phase term within the final hybrid design.

Taken together, a two-part conclusion is supported by the ablations. First, the exact memory-efficient reformulation is shown to be necessary for practicality, since naive phase-aware attention is either highly memory-intensive or impractical under the full matched setting. Second, a consistent but modest improvement in final test quality is provided by the phase term across datasets. Thus, the primary systems contribution of this work is established as the step that makes phase-aware attention usable in practice, while an additional modeling benefit is contributed by the phase component within the final hybrid design.

\begin{table*}[t]
\centering
\caption{Phase ablation for the final Q-Interference model.}
\label{tab:ablation_nophase}
\resizebox{\textwidth}{!}{\begin{tabular}{llcccc}
\hline
Dataset & Model & Best Val Loss & Test Loss & Test PPL & Peak GPU Mem (MB) \\
\hline
WikiText-103 & Q-Interference (ours) & \textbf{3.1809} & \textbf{3.1852} & \textbf{24.1718} & 4227.14 \\
WikiText-103 & Q-Interference w/o phase & 3.1830 & 3.1879 & 24.2373 & \textbf{3993.53} \\
\hline
TinyStories & Q-Interference (ours) & \textbf{1.7344} & \textbf{1.7394} & \textbf{5.6941} & 4227.14 \\
TinyStories & Q-Interference w/o phase & 1.7352 & 1.7410 & 5.7030 & \textbf{3993.53} \\
\hline
pile-10k & Q-Interference (ours) & 4.0745 & \textbf{4.7616} & \textbf{116.9275} & 4227.14 \\
pile-10k & Q-Interference w/o phase & \textbf{4.0736} & 4.7634 & 117.1429 & \textbf{3993.53} \\
\hline
small-C4 & Q-Interference (ours) & \textbf{5.5224} & \textbf{5.5090} & \textbf{246.9012} & 4227.14 \\
small-C4 & Q-Interference w/o phase & 5.5228 & 5.5093 & 246.9672 & \textbf{3993.53} \\
\hline
\end{tabular}}
\end{table*}

\section{Related Work}
\textbf{Quantum and quantum-inspired attention for language modeling}
Recent work has begun to explore whether ideas from quantum computation can enrich GPT-style attention. Early studies such as \cite{quantumSelfAttentionText2024} and \cite{quantumMixedSelfAttention2025} introduced quantum self-attention mechanisms for NLP by representing token interactions through quantum-inspired or quantum-state-based formulations. These works showed that quantum-style attention can improve representation quality, but they were evaluated mainly on text classification tasks rather than autoregressive language modeling. As a result, they do not directly address the requirements of GPT-style next-token prediction, where attention must operate efficiently over long causal contexts.
More recent work moves closer to generative GPT transformer settings. \cite{hybridTransformer2025} introduces a hybrid quantum-classical self-attention module in a decoder for sequence generation, and \cite{hybridQuantumTransformer2025} extends hybrid quantum-classical transformers to natural language generation. Most relevant to our setting, \cite{quantumInspiredSelfAttention2026} incorporates a classical quantum-inspired self-attention mechanism into the autoregressive GPT-1 pipeline. Together, these studies demonstrate that quantum or quantum-inspired attention can be used in generative transformer models. However, they mainly emphasize feasibility and architectural design, rather than the memory overhead introduced by richer attention interactions.
This limitation is especially important for quantum-inspired attention designs that introduce additional structure beyond standard dot-product similarity. Although such mechanisms can model richer token relationships, prior work has not directly addressed the specific memory overhead caused by explicitly computing phase-aware pairwise interactions. Our work focuses on this issue by reformulating the computation to avoid constructing the large intermediate interaction tensor.

\noindent
\textbf{Memory optimization in quantum-inspired LLM research}
A separate line of research uses quantum or quantum-inspired ideas for memory-efficient adaptation of large language models. \cite{quanTa2024}, \cite{liu2025a}, and \cite{raj2026quic} all aim to reduce the cost of fine-tuning by introducing parameter-efficient adaptation schemes motivated by quantum circuits or quantum parameter generation. These methods are valuable for reducing the trainable footprint of LLM adaptation, but they operate at the level of fine-tuning parameters or adapters, not at the level of attention-score computation itself. Therefore, they do not address the memory cost that arises when a richer attention rule introduces an explicit token-pair-feature interaction tensor.

More recently, \cite{qubitCache2026} studies memory reduction during inference by compressing the KV cache with a quantum-inspired probabilistic representation. This is closely related to the broader goal of efficient long-context language modeling, but it addresses a different source of memory cost. Specifically, KV-cache compression reduces the storage required for past keys and values during decoding, whereas our method targets the additional intermediate memory introduced by a naive implementation of phase-aware interference attention. In this sense, prior quantum-inspired memory-efficiency work has focused on cache compression, while our contribution lies in an exact reformulation that avoids explicitly constructing the internal interaction tensor.

Taken together, these prior studies show that quantum-inspired attention can enrich token interaction and that memory optimization has also been explored in adjacent parts of LLM systems. Our method is positioned differently: it begins from a more expressive phase-aware attention score and then derives an exact trigonometric factorization that avoids materializing the $T \times T \times d_h$ interaction tensor. We do not claim to remove the quadratic cost of dense attention altogether; rather, we specifically reduce the additional phase-specific memory overhead introduced by naive interference computation while preserving the same score exactly. To the best of our knowledge, such an exact memory-efficient factorization for phase-aware quantum-inspired attention in a GPT-style model has not been previously reported. A compact summary of representative prior work relative to Q-Interference is provided in the appendix~\ref{app:related_work}.

\section{Conclusion}

Q-Interference was introduced as a fully classical quantum-inspired attention mechanism for autoregressive language modeling, where token interactions are modeled through amplitude and learned phase. It is considered quantum-inspired because it draws on a wave-interference principle: aligned phases contribute constructively, while conflicting phases contribute destructively. This is important because standard attention mainly measures similarity, whereas phase-aware interaction provides a richer way to capture both supportive and suppressive relationships between tokens. Since a naive implementation is highly memory-intensive, an exact trigonometric factorization was derived to make the same interaction practical within a standard GPT pipeline. Across controlled experiments on benchmark datasets, Q-Interference was found to be the strongest practical model within the proposed phase-aware family, providing a favorable balance between language modeling quality and memory efficiency. The results further showed that the exact memory-efficient reformulation is essential for practicality, while the phase component contributes a small but consistent modeling benefit within the final hybrid design. Overall, these findings suggest that quantum-inspired phase interactions offer a meaningful extension of standard attention, but their usefulness depends on an exact reformulation that makes them practical at scale.

\begin{ack}
Use unnumbered first level headings for the acknowledgments. All acknowledgments
go at the end of the paper before the list of references. Moreover, you are required to declare
funding (financial activities supporting the submitted work) and competing interests (related financial activities outside the submitted work).
More information about this disclosure can be found at: \url{https://neurips.cc/Conferences/2026/PaperInformation/FundingDisclosure}.

Do {\bf not} include this section in the anonymized submission, only in the final paper. You can use the \texttt{ack} environment provided in the style file to automatically hide this section in the anonymized submission.
\end{ack}

\bibliographystyle{unsrtnat}
\bibliography{references}


\appendix

\section{Additional Related Work Summary} \label{app:related_work}
Table~\ref{tab:related_compact_appendix} provides a compact summary of representative prior work relative to Q-Interference. This table is intended as a supplementary overview of the related work discussion in the main paper, highlighting differences in GPT-style applicability, quantum-inspired design, phase-aware interaction, memory-awareness, and exact memory-efficient reformulation.

\begin{table}[h]
\centering
\small
\caption{Compact comparison of representative prior work and Q-Interference.}
\label{tab:related_compact_appendix}
\resizebox{\textwidth}{!}{\begin{tabular}{lccccc}
\hline
\textbf{Work} & \textbf{GPT-style} & \textbf{Quantum-inspired} & \textbf{Phase-aware} & \textbf{Memory-aware} & \textbf{Exact memory-efficient} \\
\hline
QSANN & $\times$ & \checkmark & $\times$ & $\times$ & $\times$ \\
QMSAN & $\times$ & \checkmark & $\times$ & $\times$ & $\times$ \\
HyQuT & \checkmark & \checkmark & $\times$ & $\times$ & $\times$ \\
QISA & \checkmark & \checkmark & $\times$ & $\times$ & $\times$ \\
QubitCache & \checkmark & \checkmark & $\times$ & \checkmark & $\times$ \\
Q-Interference (ours) & \checkmark & \checkmark & \checkmark & \checkmark & \checkmark \\
\hline
\end{tabular}}
\end{table}

\section{Proofs and Derivations for Q-Interference}
\subsection{Notation and Assumptions} \label{app:notation}

This appendix gives the algebraic details behind the Q-Interference attention reformulation. 
All derivations are written for a single attention head, since multi-head attention applies the same computation independently to each head. 
Let \(T\) denote the sequence length and let \(d_h\) denote the dimension of one attention head. 
For token \(i\) and token \(j\), we denote the nonnegative query and key amplitude vectors by \(a_i^q,a_j^k \in \mathbb{R}_{+}^{d_h}\), and the corresponding phase vectors by \(\phi_i^q,\phi_j^k \in \mathbb{R}^{d_h}\). 
The proofs below concern exact algebraic equivalence and memory accounting, not generalization performance. 
The factorized computation still produces a standard \(T \times T\) attention score matrix; it only avoids the additional \(T \times T \times d_h\) token-pair-feature tensor used by the naive phase-aware implementation.

\begin{table}[htbp]
\centering
\small
\caption{Notation used in the proof appendix.}
\label{tab:proof_notation}
\begin{tabularx}{\linewidth}{p{0.24\linewidth}X}
\toprule
\textbf{Symbol} & \textbf{Meaning} \\
\midrule
\(T\) & Sequence length. \\
\midrule
\(d_h\) & Dimension of one attention head. \\
\midrule
\(a_i^q \in \mathbb{R}_{+}^{d_h}\) & Nonnegative query amplitude vector for token \(i\). \\
\midrule
\(a_j^k \in \mathbb{R}_{+}^{d_h}\) & Nonnegative key amplitude vector for token \(j\). \\
\midrule
\(\phi_i^q \in \mathbb{R}^{d_h}\) & Query phase vector for token \(i\). \\
\midrule
\(\phi_j^k \in \mathbb{R}^{d_h}\) & Key phase vector for token \(j\). \\
\midrule
\(\widetilde{Q}^{(c)}, \widetilde{K}^{(c)}\) & Cosine-transformed query and key matrices. \\
\midrule
\(\widetilde{Q}^{(s)}, \widetilde{K}^{(s)}\) & Sine-transformed query and key matrices. \\
\midrule
\(S^{\mathrm{int}}\) & Phase-aware interference attention score matrix. \\
\bottomrule
\end{tabularx}
\end{table}

\subsection{Derivation of the Phase-Aware Interference Score} \label{app:dev_score}

We first show where the phase-aware interference score comes from. 
For each feature dimension \(r\), we represent the query and key features as complex amplitude-phase terms:
\begin{equation}
z^q_{i,r}=a^q_{i,r}e^{\mathrm{i}\phi^q_{i,r}},
\qquad
z^k_{j,r}=a^k_{j,r}e^{\mathrm{i}\phi^k_{j,r}},
\end{equation}
where \(\mathrm{i}\) is the imaginary unit and \(\overline{z^k_{j,r}}\) denotes the complex conjugate of \(z^k_{j,r}\).

\begin{proposition}
The phase-aware interference score is the scaled real part of an amplitude-phase inner product.
\end{proposition}

\begin{proof}
For a query feature \(z^q_{i,r}\) and a key feature \(z^k_{j,r}\), their conjugate product is
\begin{equation}
z^q_{i,r}\overline{z^k_{j,r}}
=
a^q_{i,r}a^k_{j,r}
e^{\mathrm{i}(\phi^q_{i,r}-\phi^k_{j,r})}.
\end{equation}
Using Euler's identity, the real part of this product is
\begin{equation}
\mathrm{Re}\left(
z^q_{i,r}\overline{z^k_{j,r}}
\right)
=
a^q_{i,r}a^k_{j,r}
\cos(\phi^q_{i,r}-\phi^k_{j,r}).
\end{equation}
Summing this real-valued interaction over all feature dimensions gives
\begin{equation}
\mathrm{Re}\left(
\sum_{r=1}^{d_h}
z^q_{i,r}\overline{z^k_{j,r}}
\right)
=
\sum_{r=1}^{d_h}
a^q_{i,r}a^k_{j,r}
\cos(\phi^q_{i,r}-\phi^k_{j,r}).
\end{equation}
After applying the standard attention scaling factor, the phase-aware interference score becomes
\begin{equation}
s^{\mathrm{int}}_{ij}
=
\frac{1}{\sqrt{d_h}}
\sum_{r=1}^{d_h}
a^q_{i,r}a^k_{j,r}
\cos(\phi^q_{i,r}-\phi^k_{j,r}).
\label{eq:appendix_interference_score}
\end{equation}
Therefore, the proposed score is exactly the scaled real part of the amplitude-phase inner product.
\end{proof}

This derivation also explains the constructive and destructive behavior of the score. 
When the phase difference \(\phi^q_{i,r}-\phi^k_{j,r}\) is close to zero, the cosine term is positive and the feature interaction is reinforced. 
When the phase difference is large, the cosine term decreases and can become negative, which suppresses the feature interaction.

\subsection{Exact Factorization of Phase-Aware Attention} \label{app:ex_factor}

We now show that the memory-efficient reformulation computes exactly the same score as the naive phase-aware interference score. 
The result follows from a direct trigonometric factorization and does not introduce any approximation.

\begin{theorem}
For any query and key amplitude-phase representations, the factorized score matrix is exactly equal to the naive phase-aware interference score matrix.
\end{theorem}

\begin{proof}
For a query token \(i\), key token \(j\), and feature dimension \(r\), the naive phase-aware interaction is
\begin{equation}
a^q_{i,r}a^k_{j,r}
\cos(\phi^q_{i,r}-\phi^k_{j,r}).
\end{equation}
Using the identity
\begin{equation}
\cos(\alpha-\beta)
=
\cos\alpha\cos\beta
+
\sin\alpha\sin\beta,
\end{equation}
we can rewrite this interaction as
\begin{equation}
a^q_{i,r}a^k_{j,r}
\cos(\phi^q_{i,r}-\phi^k_{j,r})
=
a^q_{i,r}a^k_{j,r}
\cos\phi^q_{i,r}\cos\phi^k_{j,r}
+
a^q_{i,r}a^k_{j,r}
\sin\phi^q_{i,r}\sin\phi^k_{j,r}.
\end{equation}
Define the transformed query and key components as
\begin{equation}
\widetilde{q}^{(c)}_{i,r}=a^q_{i,r}\cos\phi^q_{i,r},
\qquad
\widetilde{q}^{(s)}_{i,r}=a^q_{i,r}\sin\phi^q_{i,r},
\end{equation}
and
\begin{equation}
\widetilde{k}^{(c)}_{j,r}=a^k_{j,r}\cos\phi^k_{j,r},
\qquad
\widetilde{k}^{(s)}_{j,r}=a^k_{j,r}\sin\phi^k_{j,r}.
\end{equation}
Substituting these definitions into the interference score gives
\begin{equation}
s^{\mathrm{int}}_{ij}
=
\frac{1}{\sqrt{d_h}}
\sum_{r=1}^{d_h}
\left(
\widetilde{q}^{(c)}_{i,r}\widetilde{k}^{(c)}_{j,r}
+
\widetilde{q}^{(s)}_{i,r}\widetilde{k}^{(s)}_{j,r}
\right).
\end{equation}
This can be written as the sum of two inner products:
\begin{equation}
s^{\mathrm{int}}_{ij}
=
\frac{
\widetilde{q}^{(c)\top}_{i}\widetilde{k}^{(c)}_{j}
+
\widetilde{q}^{(s)\top}_{i}\widetilde{k}^{(s)}_{j}
}{\sqrt{d_h}}.
\label{eq:appendix_pairwise_factorized_score}
\end{equation}
Stacking all query and key vectors across the \(T\) tokens gives
\begin{equation}
S^{\mathrm{int}}
=
\frac{
\widetilde{Q}^{(c)}\widetilde{K}^{(c)\top}
+
\widetilde{Q}^{(s)}\widetilde{K}^{(s)\top}
}{\sqrt{d_h}}.
\label{eq:appendix_matrix_factorized_score}
\end{equation}
Thus, every entry of the factorized score matrix equals the corresponding naive phase-aware interference score. 
The reformulation is therefore exact and introduces no approximation.
\end{proof}

This factorization avoids explicitly constructing the additional token-pair-feature tensor required by the naive implementation. 
It does not remove the standard \(T \times T\) attention score matrix.

\subsection{Memory Cost of Naive and Factorized Computation} \label{app:mem_cost_comp}

We now compare the additional phase-specific intermediate storage required by the naive and factorized implementations. 
This analysis excludes the standard dense attention score matrix \(S^{\mathrm{int}}\in\mathbb{R}^{T\times T}\), which is still produced in both cases.

\begin{proposition}
The naive phase-aware implementation requires additional intermediate storage of order \(T^2d_h\), while the factorized implementation requires additional phase-specific storage of order \(Td_h\), excluding the standard attention score matrix.
\end{proposition}

\begin{proof}
In the naive implementation, the phase-aware interaction is materialized for every query token \(i\), key token \(j\), and feature dimension \(r\). 
This gives the intermediate tensor
\begin{equation}
\mathcal{T}_{ijr}
=
a^q_{i,r}a^k_{j,r}
\cos(\phi^q_{i,r}-\phi^k_{j,r}),
\qquad
\mathcal{T}\in\mathbb{R}^{T\times T\times d_h}.
\end{equation}
Therefore, the naive implementation requires storing \(T^2d_h\) additional interaction entries before summing over the feature dimension. 
Its additional phase-specific intermediate storage is therefore
\begin{equation}
\Theta(T^2d_h).
\end{equation}

In the factorized implementation, the computation uses the transformed matrices
\begin{equation}
\widetilde{Q}^{(c)},\widetilde{Q}^{(s)},\widetilde{K}^{(c)},\widetilde{K}^{(s)}
\in \mathbb{R}^{T\times d_h}.
\end{equation}
These four matrices require \(4Td_h\) entries in total. 
Ignoring constant factors, the additional phase-specific storage is therefore
\begin{equation}
\Theta(4Td_h)=\Theta(Td_h).
\end{equation}
The factorized computation still forms the standard attention score matrix
\begin{equation}
S^{\mathrm{int}}
=
\frac{
\widetilde{Q}^{(c)}\widetilde{K}^{(c)\top}
+
\widetilde{Q}^{(s)}\widetilde{K}^{(s)\top}
}{\sqrt{d_h}}
\in \mathbb{R}^{T\times T}.
\end{equation}
Thus, the factorization does not remove the standard dense attention matrix. 
It removes the additional \(T\times T\times d_h\) token-pair-feature tensor required by the naive phase-aware implementation.
\end{proof}

\subsection{Compatibility with Causal GPT Attention} \label{app:comp_cgpt}

We finally show that Q-Interference preserves the causal self-attention interface used in GPT-style models. 
The proposed method changes only the score function and leaves masking, normalization, value aggregation, and the remaining GPT transformer block unchanged.

\begin{lemma}
Replacing the standard dot-product score with \(S^{\mathrm{int}}\) preserves the causal attention interface.
\end{lemma}

\begin{proof}
For a sequence of length \(T\), the factorized interference computation produces a score matrix
\begin{equation}
S^{\mathrm{int}}\in\mathbb{R}^{T\times T}.
\end{equation}
This has the same shape as the standard dot-product attention score matrix. 
Let \(M\in\mathbb{R}^{T\times T}\) denote the causal mask, where entries above the causal boundary are assigned large negative values before softmax. 
Let \(V\in\mathbb{R}^{T\times d_h}\) denote the value matrix for one attention head. 
The masked attention weights are computed as
\begin{equation}
A
=
\mathrm{softmax}(S^{\mathrm{int}}+M),
\qquad
A\in\mathbb{R}^{T\times T},
\end{equation}
where the softmax is applied row-wise. 
The attention output is then
\begin{equation}
O=AV,
\qquad
O\in\mathbb{R}^{T\times d_h}.
\end{equation}
Thus, the output of the proposed attention head has the same shape as the output of a standard causal attention head. 
Therefore, residual connections, layer normalization, feed-forward blocks, and the next-token prediction objective can be used without modification.
\end{proof}

\section{Additional Main Result Analysis}

\subsection{Metric and Model Definitions} \label{app:metrix_def}

Table~\ref{tab:appendix_metric_definitions} defines the main terms used in the appendix figures. 
All quality metrics are computed for autoregressive next-token prediction. 
Peak GPU memory is reported under the fixed training setup used in the main experiments. 
These definitions clarify how to read the quality and efficiency comparisons in the following subsections.

\begin{table}[t]
\centering
\small
\caption{Definitions of metrics and model terms used in the appendix result analysis.}
\label{tab:appendix_metric_definitions}
\begin{tabularx}{\linewidth}{p{0.28\linewidth}X}
\toprule
\textbf{Term} & \textbf{Definition} \\
\midrule
Best validation loss & The lowest validation cross entropy observed during training. It is used to track model selection and training progress. Lower is better. \\
\midrule
Test loss & The cross entropy loss measured on the held-out test split after training. It measures how well the model assigns probability to the correct next token. Lower is better. \\
\midrule
Test perplexity & The exponential form of test loss. It can be read as the model's effective uncertainty when predicting the next token. Lower is better. \\
\midrule
Peak training GPU memory & The maximum GPU memory used during training under the reported batch size, context length, precision, and hardware. It measures practical training cost rather than language quality. Lower is better. \\
\midrule
Memory reduction & The percentage decrease in peak GPU memory relative to a chosen baseline. A larger reduction means the model needs less GPU memory under the same reported setting. \\
\midrule
Parameter matched comparison & A comparison where models have similar parameter counts. This helps reduce the chance that differences are caused only by model size. \\
\midrule
Internal custom model & A model trained in our controlled experimental pipeline. This includes the standard GPT baseline, Q-GPT baseline, naive interference model, and Q-Interference. \\
\midrule
Pre-trained reference model & A pretrained model used only for context. GPT-Neo-125M and OPT-125M are not treated as parameter matched internal baselines. \\
\midrule
Naive interference model & A direct phase-aware implementation that materializes the token-pair-feature interaction tensor. It is used to show the memory cost without the exact efficiency. \\
\midrule
Q-Interference & The proposed phase-aware model with exact memory-efficient factorization. It computes the same interference score while avoiding the large interaction tensor. \\
\bottomrule
\end{tabularx}
\end{table}

\subsection{Controlled WikiText-103 Result} \label{app:wiki_result}

WikiText-103 is the main controlled dataset in our experiments. 
Figure~\ref{fig:wikitext103_controlled_appendix} summarizes the internal comparison using two metric rulers. 
The top ruler reports test perplexity and the bottom ruler reports peak training GPU memory. 
Lower values are better for both metrics. 
Q-Interference obtains the lowest test perplexity among the internal models on WikiText-103. 
It also reduces peak GPU memory from 8055.76 MB for the standard GPT baseline to 4227.14 MB, which corresponds to an approximate 47.5\% reduction while preserving the same GPT-style training objective.

\begin{figure}[htbp]
    \centering
    \includegraphics[width=0.98\linewidth]{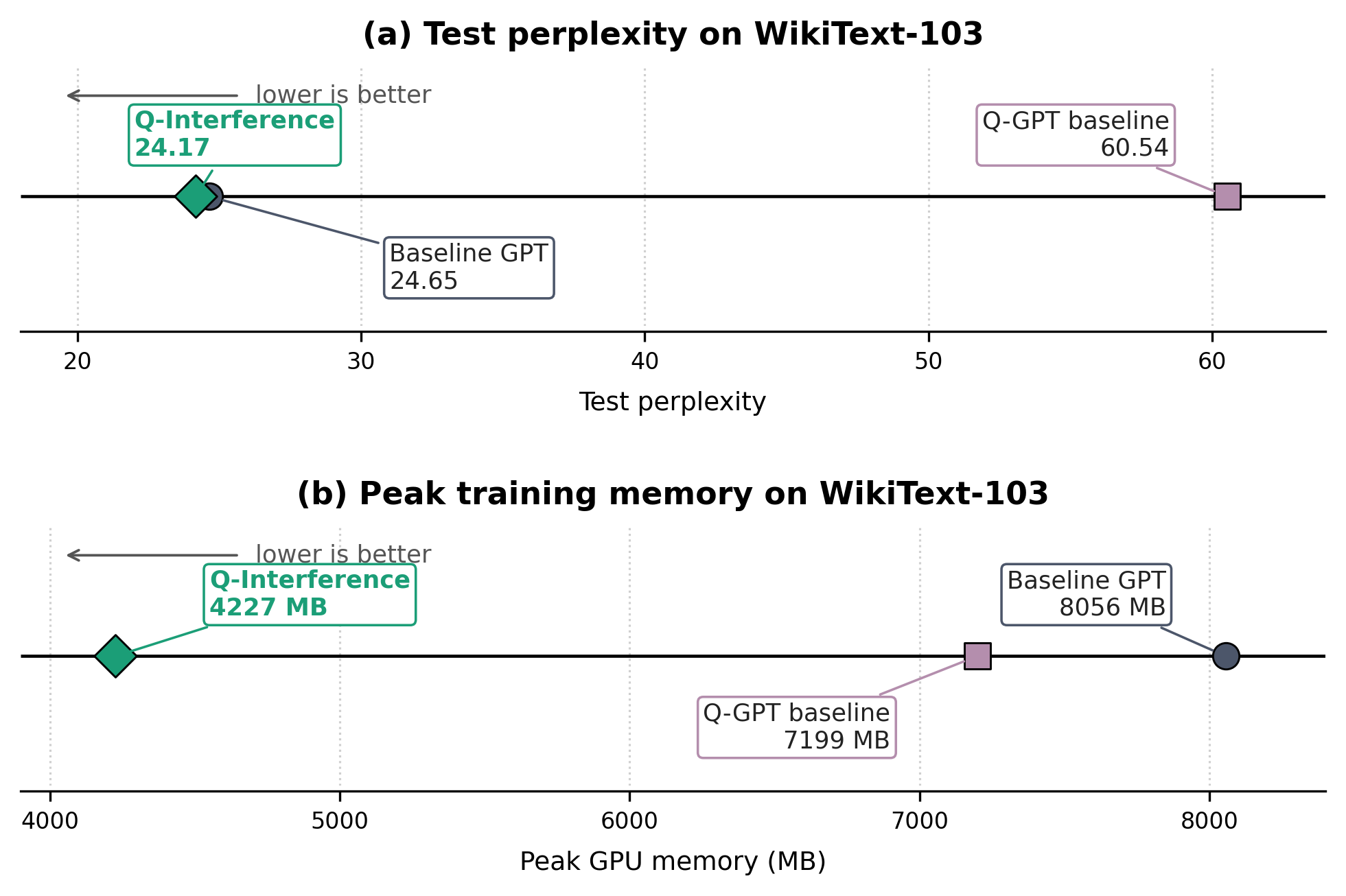}
    \caption{Controlled comparison on the WikiText-103 dataset. The top ruler reports test perplexity and the bottom ruler reports peak training GPU memory. Each marker represents one internal model, and each label reports the exact value. Q-Interference achieves the best internal test perplexity and the lowest peak GPU memory in this controlled setting.}
    \label{fig:wikitext103_controlled_appendix}
\end{figure}

\subsection{Cross-Dataset Memory Behavior} \label{app:data_memory}

Figure~\ref{fig:cross_dataset_memory_appendix} shows the peak training GPU memory of the custom model family on TinyStories, pile-10k, and small-C4. 
The naive interference model uses the largest memory because it directly materializes the phase-aware token-pair-feature interaction tensor. 
In contrast, Q-Interference avoids this tensor through the exact factorized computation. 
As a result, Q-Interference has the lowest peak memory among the custom models in all three cross-dataset settings. 
The repeated memory values across datasets reflect the fixed profiling setup, including batch size, context length, numerical precision, and hardware.

\begin{figure}[t]
    \centering
    \includegraphics[width=0.92\linewidth]{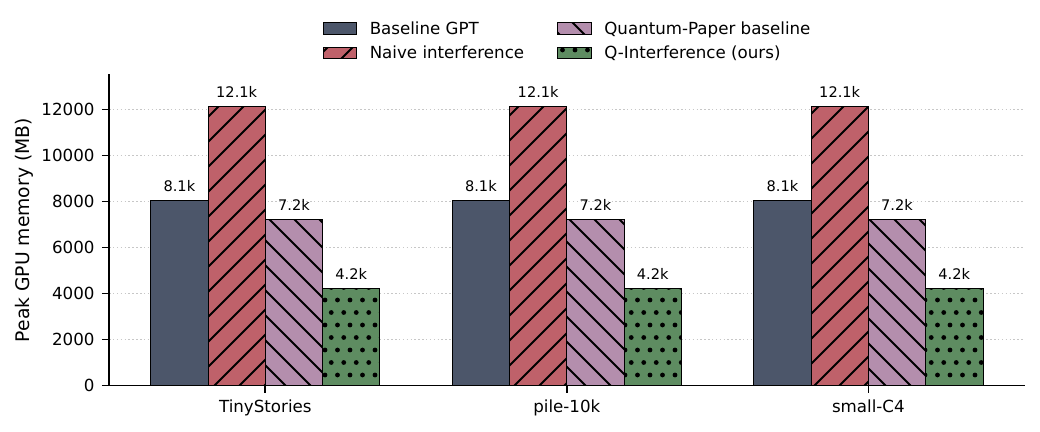}
    \caption{Peak GPU memory across the custom model family. Q-Interference consistently uses less memory than the standard GPT baseline, naive interference, and the Q-GPT baseline under the reported setting. The naive interference model highlights the cost of materializing the phase-aware interaction tensor.}
    \label{fig:cross_dataset_memory_appendix}
\end{figure}

\subsection{Pretrained 125M  Models} \label{app:pretrain}

Figure~\ref{fig:external_reference_appendix} compares Q-Interference with GPT-Neo-125M and OPT-125M. 
These two models are pretrained decoder-only language models, while Q-Interference is trained from scratch in our controlled experimental setup. 
Therefore, this comparison should be read as contextual evidence rather than as a parameter-matched baseline comparison. 
The pretrained models achieve stronger final test perplexity, which is expected because they benefit from large-scale pretraining. 
However, Q-Interference remains competitive in training memory and uses lower peak GPU memory than both pretrained references on TinyStories, pile-10k, and small-C4. 
This result suggests that the proposed phase-aware attention design can be implemented practically, even when compared against strong pretrained 125M-scale reference models.

\begin{figure}[t]
    \centering
    \includegraphics[width=0.94\linewidth]{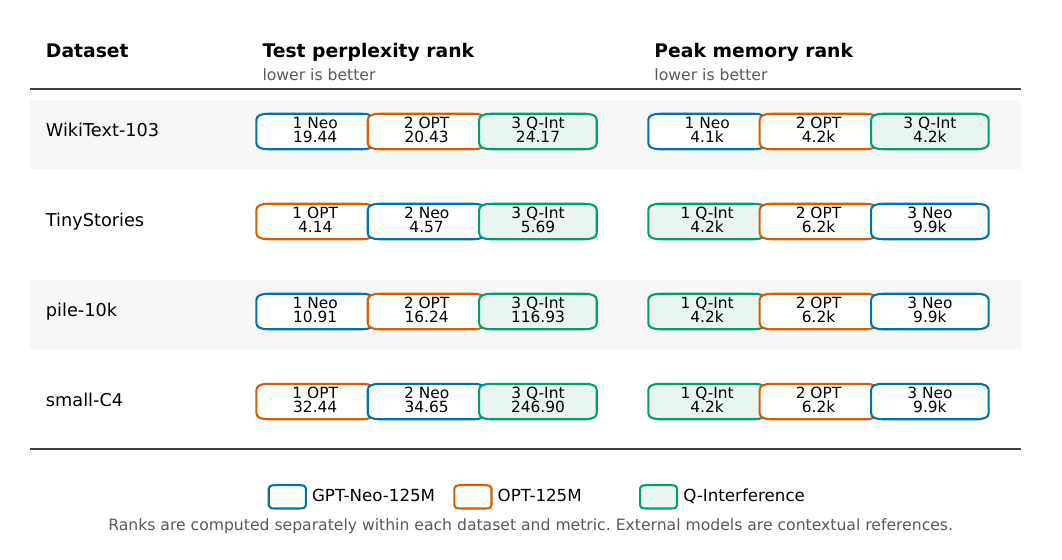}
    \caption{Pretrained 125M  model comparison. GPT-Neo-125M and OPT-125M are pretrained decoder-only models, while Q-Interference is trained from scratch in the controlled setup used in this work. The pretrained references obtain stronger final language-modeling quality, but Q-Interference shows a favorable memory profile and uses lower peak GPU memory on three of the four datasets. The comparison is intended to contextualize Q-Interference rather than to serve as a parameter-matched baseline.}
    \label{fig:external_reference_appendix}
\end{figure}

\section{Limitations and Discussion} \label{app:limit_discuss}
The results support Q-Interference as a practical and well-motivated phase-aware attention design. The strongest evidence in this work is that the proposed exact reformulation makes richer interference-based attention trainable in a standard GPT pipeline while substantially reducing the additional memory cost of naive phase-aware computation. The ablations further show that the phase term contributes a small but consistent modeling benefit on top of this practical design.

At the same time, the method should be interpreted with appropriate scope. Q-Interference is not intended as a full replacement for all strong pretrained GPT transformer baselines, and the current study focuses on controlled GPT-style language modeling rather than large-scale pretraining or downstream transfer. Even so, the results suggest that phase-aware quantum-inspired interactions are worth studying as a richer attention family, especially when paired with an exact memory-efficient reformulation that makes them practical.


\clearpage
\input{checklist.tex}

\end{document}

%% file: checklist.tex
\section*{NeurIPS Paper Checklist}



\begin{enumerate}

\item {\bf Claims}
    \item[] Question: Do the main claims made in the abstract and introduction accurately reflect the paper's contributions and scope?
    \item[] Answer: \answerYes{} 
    \item[] Justification: The main claims are stated in the abstract and Introduction.

\item {\bf Limitations}
    \item[] Question: Does the paper discuss the limitations of the work performed by the authors?
    \item[] Answer: \answerYes{} 
    \item[] Justification: Limitations are discussed in the Appendix D at page 17.

\item {\bf Theory assumptions and proofs}
    \item[] Question: For each theoretical result, does the paper provide the full set of assumptions and a complete (and correct) proof?
    \item[] Answer: \answerYes{} 
    \item[] Justification: The assumptions and derivations are provided in Appendix A. The appendix proves the phase-aware score, exact factorization, memory accounting and compatibility with causal GPT attention.

    \item {\bf Experimental result reproducibility}
    \item[] Question: Does the paper fully disclose all the information needed to reproduce the main experimental results of the paper to the extent that it affects the main claims and/or conclusions of the paper (regardless of whether the code and data are provided or not)?
    \item[] Answer: \answerYes{} 
    \item[] Justification: The current submission describes the experimental setup.

\item {\bf Open access to data and code}
    \item[] Question: Does the paper provide open access to the data and code, with sufficient instructions to faithfully reproduce the main experimental results, as described in supplemental material?
    \item[] Answer: \answerYes{} 
    \item[] Justification: All dataset are public. We have provided our project git link in the abstract. 

\item {\bf Experimental setting/details}
    \item[] Question: Does the paper specify all the training and test details (e.g., data splits, hyperparameters, how they were chosen, type of optimizer) necessary to understand the results?
    \item[] Answer: \answerNo{} 
    \item[] Justification: The paper reports the datasets, model families, context length, hardware, precision, and evaluation metrics in Section 3. However, it does not explicitly specify all training details such as optimizer type, learning rate, batch size, number of epochs, seed choice, and hyperparameter selection procedure.

\item {\bf Experiment statistical significance}
    \item[] Question: Does the paper report error bars suitably and correctly defined or other appropriate information about the statistical significance of the experiments?
    \item[] Answer: \answerNo{} 
    \item[] Justification: The paper reports single-run values for validation loss, test loss, test perplexity, and peak GPU memory, but it does not report error bars, confidence intervals, statistical significance tests, or multi-seed mean ± standard deviation for the main experimental results.

\item {\bf Experiments compute resources}
    \item[] Question: For each experiment, does the paper provide sufficient information on the computer resources (type of compute workers, memory, time of execution) needed to reproduce the experiments?
    \item[] Answer: \answerYes{} 
    \item[] Justification: We provide all key information in our submission.
    
\item {\bf Code of ethics}
    \item[] Question: Does the research conducted in the paper conform, in every respect, with the NeurIPS Code of Ethics \url{https://neurips.cc/public/EthicsGuidelines}?
    \item[] Answer: \answerYes{} 
    \item[] Justification: Yes, we maintain all the ethics related to code.

\item {\bf Broader impacts}
    \item[] Question: Does the paper discuss both potential positive societal impacts and negative societal impacts of the work performed?
    \item[] Answer: \answerNA{} 
    \item[] Justification: There is no ethical consideration right now.
    
\item {\bf Safeguards}
    \item[] Question: Does the paper describe safeguards that have been put in place for responsible release of data or models that have a high risk for misuse (e.g., pre-trained language models, image generators, or scraped datasets)?
    \item[] Answer: \answerNA{} 
    \item[] Justification: This paper does not contain such resources.

\item {\bf Licenses for existing assets}
    \item[] Question: Are the creators or original owners of assets (e.g., code, data, models), used in the paper, properly credited and are the license and terms of use explicitly mentioned and properly respected?
    \item[] Answer: \answerYes{} 
    \item[] Justification: Yes, we have cited properly.

\item {\bf New assets}
    \item[] Question: Are new assets introduced in the paper well documented and is the documentation provided alongside the assets?
    \item[] Answer: \answerNA{} 
    \item[] Justification: There is no new asset. 

\item {\bf Crowdsourcing and research with human subjects}
    \item[] Question: For crowdsourcing experiments and research with human subjects, does the paper include the full text of instructions given to participants and screenshots, if applicable, as well as details about compensation (if any)? 
    \item[] Answer: \answerNA{} 
    \item[] Justification: No crowdsourcing is in here.

\item {\bf Institutional review board (IRB) approvals or equivalent for research with human subjects}
    \item[] Question: Does the paper describe potential risks incurred by study participants, whether such risks were disclosed to the subjects, and whether Institutional Review Board (IRB) approvals (or an equivalent approval/review based on the requirements of your country or institution) were obtained?
    \item[] Answer: \answerNA{} 
    \item[] Justification: This paper doesn't include human subjects. 

\item {\bf Declaration of LLM usage}
    \item[] Question: Does the paper describe the usage of LLMs if it is an important, original, or non-standard component of the core methods in this research? Note that if the LLM is used only for writing, editing, or formatting purposes and does \emph{not} impact the core methodology, scientific rigor, or originality of the research, declaration is not required.
    \item[] Answer: \answerNA{} 
    \item[] Justification: We used LLM for editing, grammar checking and word choice. 

\end{enumerate}